\documentclass[10pt, a4paper]{article}
\usepackage{lrec2000}

\title{Bootstrapping a Tagged Corpus through Combination of Existing
	Heterogeneous Taggers}

\name{Jakub Zavrel, Walter Daelemans}

\address{CNTS / Language Technology Group\\
	 University of Antwerp\\
	 Universiteitsplein 1\\ 
	 2610 Wilrijk, Belgium\\
	 \{zavrel,daelem\}@uia.ua.ac.be}

\abstract{This paper describes a new method, {\sc combi-bootstrap}, to
exploit existing taggers and lexical resources for the annotation of
corpora with new tagsets. {\sc combi-bootstrap} uses existing
resources as features for a second level machine learning module, that
is trained to make the mapping to the new tagset on a very small
sample of annotated corpus material. Experiments show that {\sc
combi-bootstrap}: i) can integrate a wide variety of existing
resources, and ii) achieves much higher accuracy (up to 44.7 \% error
reduction) than both the best single tagger and an ensemble tagger
constructed out of the same small training sample.}

\begin{document}

\maketitleabstract

\section{Introduction}
When morpho-syntactically annotating a corpus with a new tagset, the
initial stages of the annotation process face a bootstrapping problem.
There are no automatic taggers available to help the annotator, and
because of this, the annotation process is too laborious to quickly
produce adequate amounts of training material for the tagger. A
solution which has been suggested in previous
work~\cite{Teufel+95,Atwell+94}, is to use an existing tagger, and
devise mapping rules between the old and the new tagset. However, as
the construction of such mapping rules requires considerable
linguistic knowledge engineering, this solution only shifts the
problem to a different domain.

In this paper we describe a new method that uses machine learning and
a very small corpus sample annotated in the new tagset.  It allows us
to exploit existing taggers and lexical resources with a wild
variation in tagsets to quickly reach a level of tagging accuracy far
beyond that of taggers trained on the initially very small annotated
samples.

The idea behind this method, which we will refer to as {\sc
combi-bootstrap}, comes from previous work on combining taggers to
improve accuracy~\cite{VanHalteren+98,VanHalteren+00,Brill+98}. These
approaches combine a number of taggers, all trained on the same corpus
data and using the same tagset, to yield a combined tagger that has a
much higher accuracy than the best component system. The reasoning
behind this is that the components make different errors, and a
combination method is able to exploit these differences. Simple
combination methods, such as (weighted) voting, are confined to output
that is i) in the same tagset as the components, and ii) is one of the
tags suggested by the components. However, more sophisticated
combination methods exist, which do not share these limitations. In
{\em Stacking}~\cite{Wolpert92b}, the outputs of the component systems
are used as features for a second level machine learning module, that
is trained on held out data to correct the errors that the components
make. First, this theoretically allows the second level learner to
recognize situations where {\em all} components are in error, and
correct these. Second, this lifts the requirement that the components
use the same vocabulary of categories. We can in effect present the
second level learner with any type of representations of the context
to be tagged, such as the word itself, but also output from existing
taggers with other tagsets. The positive effects of this approach are
demonstrated in the remainder of this paper. This is structured
as follows. In Section~\ref{Data} we describe the data sets that are
used in the experiments. In Section~\ref{Systems} we describe the
component taggers and the machine learning method used for the second
level learner. In Section~\ref{Results} we present the results of our
experiments using a variety of combination setups. And finally, in
Section~\ref{Conclusion}, we summarize and conclude.

\section{Data}
\label{Data}

We developed and tested our bootstrapping method in the context of the
morpho-syntactic annotation of the ``Corpus Gesproken Nederlands''
(Spoken Dutch Corpus; henceforth called CGN)~\cite{VanEynde+00a}. For
this corpus, a fine-grained tagset was developed that distinguishes
morphological and syntactic features such as number, case, tense,
etc. for a total of approximately 300 tags. Annotation of this corpus
has only just started, so we conducted experiments on three small
samples (of respectively 5, 10 and 20 thousand tokens, including
punctuation) of the initial corpus~\footnote{These were annotated by
manually correcting tags produced by the first {\sc combi-bootstrap}
taggers}.

As existing Dutch resources we use four popular taggers (described in
Section~\ref{Systems}) trained on (parts of) the written sections of
the Eindhoven corpus~\cite{UitdenBoogaart75}, tagged with either the
{\sc wotan-1} (347 tags) or {\sc wotan-lite} (both with 641424 tokens of
training data) or {\sc wotan-2} (1256 tags, and a slightly more modest
126803 tokens of training data) ~\cite{Berghmans94,VanHalteren99b}
tagsets.  Furthermore we will use the ambiguous lexical
categories\footnote{Not including function words like determiners
pronouns etc. I.e.~adjective, adverb, noun, number, exclamation,
verb.} of words taken from the CELEX~\cite{Baayen+93} lexical
database. The section of this database that we use, contains
300837 distinct word forms.


On this data we measure the accuracy of single taggers trained on 90\%
of the data and tested on the remaining 10\%. To test the accuracy of
a combined system, the 90\% training data is split into nine pieces,
and the four component taggers are tested on each part in turn (and
trained on the remaining eight pieces, i.e.~nine-fold
cross-validation). The test outputs of the taggers on the nine
training pieces are then concatenated and used as training material
for the second level combination learner, which is tested on the
reserved 10\% test material. When examining the effects of including
existing resources in the combination, both train and test set are
tagged using some tagging system (e.g. an HMM tagger using {\sc
wotan-1}, or the ambiguous lexical categories from CELEX), and the
effect is measured as the accuracy of the second level learner in
predicting the target CGN tagging 
for the test set.

\section{Systems}
\label{Systems}

We experimented with four well-known trainable part of speech
taggers: TNT (a trigram HMM tagger~\cite{Brants00}), MXPOST (A Maximum
Entropy tagger;~\cite{Ratnaparkhi96}, henceforth referred to as MAX),
The~\cite{Brill95} Rule based tagger (referred to as RUL), and MBT (a
Memory-Based tagger;~\cite{Daelemans+96b}). The RUL tagger was not
available trained on the {\sc wotan} resources, because its training
is too expensive on large corpora with large tagsets.

As the combination method we have used IB1~\cite{Aha+91} a Memory Based
Learning method implemented in the TiMBL\footnote{Available from {\tt
http://ilk.kub.nl/}}~\cite{Daelemans+00} system. IB1 stores the
training set in memory and classifies test examples by returning the
most frequent category in the set of $k$ nearest neighbors (i.e. the
least distant training patterns). In the experiments below, we use the
Overlap distance metric, no feature weighting, and $k=1$.

\section{Results}
\label{Results}

\subsection{Baselines}

When we train the separate taggers on training sets from the CGN
corpus of three consecutive sizes, we obtain the accuracies shown
Table~\ref{baseline}. We also show the percentage of unknown words in
each of the test partitions. Unknown words are defined as tokens that
are not found in the 90\% training partition. From this we can see
that the performance on unknown words is a major component of the
bootstrapping problem. We see that TNT has the best overall score for
all three training set sizes (resp. 84.49, 86.39, and 90.75 \%
correct). It also has the best scores for known words. Only for
unknown words does it find a serious contender in MAX. When we do a
straightforward combination of the four taggers in the style
of~\cite{VanHalteren+00} with IB1 as the second level learner we get a
combined tagger with an accuracy of resp. 84.32, 87.24 and 90.46 \%
correct for the 5k, 10k and 20k data sets. Only for the 10k set this
is better than the best individual tagger. The reason we do not obtain
accuracy gains as in~\newcite{VanHalteren+98} here, is probably that
the number of training cases for the second level learner is too small
at this data set size. Also, as was shown in~\newcite{VanHalteren+98},
IB1 is not the best combiner at small training set sizes. However, to
keep the comparison simple, we will not use weighted voting
combination here (which does perform better at small training set
sizes), because voting approaches cannot be used for the {\sc
combi-bootstrap} method.

\begin{table*}[t]
\begin{center}
\begin{tabular}{|l|rrr|rrr|rrr|}
\hline
    & \multicolumn{9}{|c|}{Data set size}\\
\hline
tagger &\multicolumn{3}{|c|}{5000} & \multicolumn{3}{|c|}{10000} & \multicolumn{3}{|c|}{20000}\\
       & u & k & t & u & k & t & u & k & t\\
\hline
MBT    & 39.42 & 90.84 & 82.01 & 46.25 & 91.57 & 85.36 & 45.93 & 93.03 & 88.29 \\
TNT    & 49.04 & 91.83 & 84.49 & 50.00 & 92.16 & 86.39 & 57.42 & 94.48 & 90.75 \\
MAX    & 50.00 & 79.48 & 74.42 & 58.13 & 86.21 & 82.36 & 57.42 & 90.35 & 87.04 \\
RUL    & 29.81 & 87.65 & 77.72 & 37.50 & 87.50 & 80.65 & 40.19 & 89.71 & 84.72 \\
\hline
CGN ensemble & & & 84.32 & & & 87.24 & & & 90.46\\ 
\hline
\% unknown &  &   & 17.16  &  &   & 13.69  &  &   & 10.07 \\
\hline
\end{tabular}
\caption{Test set accuracies for taggers trained on 90\% of the CGN
data and tested on 10\%. The accuracies for the single taggers are
given separately for unknown (u), known (k), and all (t) tokens. The
bottom row gives the percentage of unknown words for the test
partition.}
\label{baseline}
\end{center}
\end{table*}

\subsection{{\sc Combi-bootstrap}: Reusing existing resources}

In this section we will add, one by one, a number of resources that
use different tagsets. In contrast to the native CGN taggers, these
resources have much larger lexical coverage, and the taggers among
them have been trained on much larger corpora (see data description in
Section~\ref{Data}). We will call the resources: CGN, for the block of
four CGN-taggers trained in the previous section, {\sc Word} for the
word to be tagged itself, CEL for the ambiguous categories on the
basis of CELEX. W1, W2, and WL stand for {\sc wotan} 1, 2 and Lite
blocks respectively (each of which contains three different taggers:
MBT, MAX, and TNT). And, finally, {\sc Wall} stands for the set of all
(nine) {\sc wotan}-based taggers. The way the resources are added is
by including them as features in the case representation for the
second level learner. Figure~\ref{example} illustrates this
representation for the case of all sources being used.

\begin{figure*}
\begin{center}
\begin{tiny}
\begin{tabular}{|l|p{0.75cm}p{0.75cm}p{0.75cm}p{0.75cm}|p{0.75cm}|p{0.75cm}p{0.75cm}p{0.75cm}|p{0.75cm}p{0.75cm}p{0.75cm}|p{0.75cm}p{0.75cm}p{0.75cm}|p{0.75cm}|}
\hline
{\sc Word} & \multicolumn{4}{|c|}{CGN} & CEL & \multicolumn{3}{|c|}{W1} & \multicolumn{3}{|c|}{W2} & \multicolumn{3}{|c|}{WL} & CGN Target\\
           & MAX & MBT & RUL & TNT &  & MAX & MBT & TNT & MAX & MBT & TNT & MAX & MBT & TNT & \\
\hline
omdat & VG(onder) & VG(onder) & VG(onder) & VG(onder) & UNKNOWN & Conj(onder, metfin) & Conj(onder, metfin) & Conj(onder, metfin) & Conj(subord, withfin, conj) & Conj(subord, withfin, conj) & Conj(subord, withfin, conj) & Conj(onder, metfin) & Conj(onder, metfin) & Conj(onder, metfin) & VG(onder)\\
\hline
ik & VNW(pers, pron, nomin, vol, 1, ev) & VNW(pers, pron, nomin, vol, 1, ev) & VNW(pers, pron, nomin, vol, 1, ev) & VNW(pers, pron, nomin, vol, 1, ev) & substantief & Pron(per, 1, ev, nom) & Pron(per, 1, ev, nom) & Pron(per, 1, ev, nom) & Pron(pers, first, sing, nom, str, nom) & Pron(pers, first, sing, nom, str, nom) & Pron(pers, first, sing, nom, str, nom) & Pron(per, 1, ev, nom) & Pron(per, 1, ev, nom) & Pron(per, 1, ev, nom) & VNW(pers, pron, nomin, vol, 1, ev)\\
\hline
voor & VZ(init) & VZ(init) & VZ(init) & VZ(init) & substantief & Prep(voor) & Prep(voor) & Prep(voor) & Adp(prep, obl\^{ }obl + dat, adp\^{ }adp + nampart) & Adp(prep, obl\^{ }obl + dat, adp\^{ }adp + nampart) & Adp(prep, obl\^{ }obl + dat, adp\^{ }adp + nampart) & Prep & Prep & Prep & VZ(init)\\
\hline
de & LID(bep, stan, rest) & LID(bep, stan, rest) & LID(bep, stan, rest) & LID(bep, stan, rest) & UNKNOWN & Art(bep, zijdofmv, neut) & Art(bep, zijdofmv, neut) & Art(bep, zijdofmv, neut) & Art(def, nonsingn, det\^{ }det + nampart) & Art(def, nonsingn, det\^{ }det + nampart) & Art(def, nonsingn, det\^{ }det + nampart) & Art(bep, zijdofmv, neut) & Art(bep, zijdofmv, neut) & Art(bep, zijdofmv, neut) & LID(bep, stan, rest)\\
\hline
klas & N(soort, ev, basis, zijd, stan) & N(soort, ev, basis, zijd, stan) & N(soort, ev, basis, zijd, stan) & N(soort, ev, basis, zijd, stan) & substantief & N(soort, ev, neut) & N(soort, ev, neut) & N(soort, ev, neut) & N(com, singmf, nom) & N(com, singmf, nom) & N(com, singmf, nom) & N(ev, neut) & N(ev, neut) & N(ev, neut) & N(soort, ev, basis, zijd, stan)\\
\hline
sta & BW() & WW(pv, tgw, ev) & WW(pv, tgw, ev) & WW(pv, tgw, ev) & werkwoord & V(intrans, ott, 1, ev) & V(intrans, ott, 2, ev) & V(intrans, ott, 1, ev) & N(prop, sing, nom\^{ }nom + nampart) & V(lex, intrans\^{ }intrans + trans, pres, s1\^{ }s1 + s2i, hebben, nonsep, verb) & V(lex, intrans\^{ }intrans + trans, pres, s1\^{ } s1 + s2i, hebben, nonsep, verb) & V(ott, 1, ev) & V(ott, 1, ev) & V(ott, 2, ev) & WW(pv, tgw, ev)\\
\hline
\end{tabular}
\end{tiny}
\caption{An example of case representations for the second level learner with all information sources as features.}
\label{example}
\end{center}
\end{figure*}

\begin{table}
\begin{center}
\begin{tabular}{|l|rrr|}
\hline
	         & \multicolumn{3}{|c|}{Data set size}\\
	         & 5000 & 10000 & 20000\\
\hline
CGN		 & 84.32 & 87.24 & 90.46\\
\hline
CGN +{\sc Word}  & 83.66 & 87.59 & 90.46\\
CGN + CEL        & 85.64 & 88.18 & 91.18\\
CGN + W1	 & 89.11 & 90.50 & 92.39\\
CGN + WL	 & 88.45 & 90.24 & 92.48\\
CGN + W2	 & 88.94 & 89.55 & 91.61\\
\hline
\end{tabular}
\caption{The effect of adding existing information sources one by one.}
\label{CGNplusone}
\end{center}
\end{table}

First we consider the effects of adding the information sources one by
one to CGN. The results are shown in Table~\ref{CGNplusone}. This
shows that every added resource has a positive effect. The largest
improvement is obtained by adding the {\sc wotan} taggers.  Second, we
tried to leave out the CGN block all together, and test the value of
{\em only} the other information sources. This results in the scores
shown in Table~\ref{noCGN}. Interestingly, we see that the separate
existing resources by themselves are not very good predictors at
all. In particular CELEX (with only ambiguous main parts of speech)
scores poorly. But also the blocks of three {\sc wotan} taggers (MAX,
TNT, MBT) with the same tagset (either W1, W2 or WL) are worse than
the best CGN taggers trained from scratch. However, this is changed
when we use the {\sc Wall} combination: all 3 (algorithms) times 3
(tagsets) {\sc wotan} taggers. In fact, this block, together with
CELEX and the word itself, performs better (92.82\% at 20k) than
the best CGN+{\sc wotan} combination so far (92.48\%). These results
also show that CELEX and {\sc Word} are valuable additions, even
though they are poor predictors by themselves.

\begin{table}
\begin{center}
\begin{tabular}{|l|rrr|}
\hline
	         & \multicolumn{3}{|c|}{Data set size}\\
	         & 5000 & 10000 & 20000\\
\hline
{\sc Word}	 & 73.10 & 75.60 & 80.05\\
CEL		 & 25.74 & 27.40 & 29.49\\
W1		 & 81.35 & 82.45 & 82.65\\
WL		 & 78.38 & 77.31 & 77.35\\
W2		 & 83.83 & 86.64 & 86.89\\
{\sc Wall}	 & 90.10 & 91.01 & 91.47\\
{\sc Wall} + CEL & 90.10 & 91.01 & 91.47\\
{\sc Wall} + {\sc Word} & 90.92 & 91.52 & 92.43\\
{\sc Wall} + CEL + {\sc Word} & 91.25 & 91.52 & 92.82\\
\hline
\end{tabular}
\caption{The effect of the information sources without the contribution of the CGN block.}
\label{noCGN}
\end{center}
\end{table}

Finally, we threw all the information sources together in the
combiner. This has a further positive effect, as can be seen in
Table~\ref{AllNow}. In fact, it seems that more sources is simply
better~\footnote{We have, however, not tried to check this
exhaustively by leaving out single CGN or {\sc wotan} taggers.}. The
best result (93.49\% correct with all information sources at 20k data
set size) shows 2.74\% less errors than the best single CGN tagger, a
29.6\% error reduction. The error reduction is even larger for smaller
data set sizes, as can be seen in Table~\ref{errorreduct}. In this
table, the error reduction is also shown separately for known and
unknown words. The gain for unknown words is dramatically larger than
that for known words, showing that the effect of our method can mostly
be attributed to the larger lexical coverage of the existing
resources. Further analysis would be needed to separate this from the
effect of better ``unknown word guessing'' of the existing taggers.

\begin{table}
\begin{center}
\begin{tabular}{|l|rrr|}
\hline
	         & \multicolumn{3}{|c|}{Data set size}\\
	         & 5000 & 10000 & 20000\\
\hline
CGN + {\sc Wall} 			& 91.25 & 91.44 & 93.40\\
CGN + {\sc Wall} + {\sc Word} 		& 91.42 & 91.44 & 93.35\\
CGN + {\sc Wall} + CEL 			& 91.25 & {\bf 91.78} & 93.45\\ 
CGN + {\sc Wall} + CEL + {\sc Word} 	& {\bf 91.42} & 91.70 & {\bf 93.49}\\
\hline
\end{tabular}
\caption{The effect of large combinations. The boldface figures
indicate the best results overall from this paper.}
\label{AllNow}
\end{center}
\end{table}

\begin{table*}
\begin{center}
\begin{tabular}{|l|rrr|rrr|rrr|}
\hline
    & \multicolumn{9}{|c|}{Data set size}\\
\hline
tagger &\multicolumn{3}{|c|}{5000} & \multicolumn{3}{|c|}{10000} & \multicolumn{3}{|c|}{20000}\\
       & u & k & t & u & k & t & u & k & t\\
\hline
best single CGN (TNT)  & 49.04 & 91.83 & 84.49 & 50.00 & 92.16 & 86.39 & 57.42 & 94.48 & 90.75 \\
best {\sc combi-bootstrap} & 75.00 & 94.82 & 91.42 & 78.13 & 93.45 & 91.70 & 76.08 & 95.44 & 93.49\\
\hline
$\Delta$ error (\%) & -50.9 & -36.6 & -44.7 & -56.3 & -16.5 & -39.0 & -43.8 & -17.4 & -29.6\\
\hline
\end{tabular}
\caption{Accuracy of the best {\sc combi-bootstrap} system (the one using
all information sources) and the best individual tagger trained only
on the CGN data, and the associated percentage of error reduction. The scores are split out into unknown (u) and known (k) words, and total (t).}
\label{errorreduct}
\end{center}
\end{table*}

Because the combination of all information sources contains sources of
a very diverse character, a plausible intuition would be that feature
weighting could help the Memory-Based classifier. However, further
experimentation with TiMBL parameters showed that no parameter setting
had a significant gain over unweighted Overlap with $k=1$ for this
data set. This would probably be different if we had more data to
train the combiner on. However, such luxury is not typical of the main
application context of the proposed method.

\section{Conclusion}
\label{Conclusion}

We have described {\sc combi-bootstrap}, a new method for
bootstrapping the annotation of a corpus with a new tagset from
existing information sources in the same language and very small
samples of hand-annotated material. {\sc Combi-bootstrap} is based on
the principle of Stacking machine learning algorithms, and shows very
good performance on the CGN corpus that we have experimented with.
The best performance was obtained when all available information
sources are used at the same time, which yields an error reduction of
up to 44.7\% in one case. As the test samples are very small, however,
further experimentation will be needed on other corpora.

Most importantly, we have shown that if existing resources are
available, a tagger for a new corpus and tagset can quickly be lifted
into a workable accuracy-range for manual correction. Moreover, the
proposed method seems promising for application in other domains such
as word sense disambiguation or parsing, where large training resouces
are difficult to construct and existing representation schemes are
very diverse.

\section*{Acknowledgements}

Parts of this research were supported by the project ``Spoken Dutch
Corpus (CGN)'' which is funded by the Netherlands Organization for
Scientific Research (NWO) and the Flemish Government.  We would like
to thank Hans van Halteren, and the CGN Corpus Annotation Working
Group for respectively the availability of the {\sc wotan} data sets
and CGN corpus samples. Furthermore we wish to thank Antal van den
Bosch for stimulating discussions concerning this research. The
authors of this research are (partially) supported by a grant from the
Centre for Evolutionary Language Engineering, Flanders Language Valley
S.AI.L Port.

\bibliographystyle{lrec2000}
\bibliography{cgn-combi}

\end{document}